\pdfoutput=1
\documentclass[11pt]{article}

\usepackage{ACL2023}

\usepackage{times}
\usepackage{latexsym}

\usepackage[T1]{fontenc}
\usepackage[utf8]{inputenc}

\usepackage{microtype}
\usepackage{physics}

\usepackage{inconsolata}
\usepackage{tabularx}
\usepackage{multirow}
\usepackage{makecell}
\usepackage{booktabs}
\usepackage{enumitem}

\usepackage{stfloats}
\usepackage{placeins}
\usepackage{caption}
\usepackage{float}

\renewcommand{\arraystretch}{1.3}

\title{Enhancing Scientific Visual Question Answering through Multimodal Reasoning and Ensemble Modeling}

\author{\textbf{Prahitha Movva} \\
  University of Massachusetts Amherst \\
  Amherst, MA, USA \\
  \texttt{prahitha.movva03@gmail.com} \\\And
  Naga Harshita Marupaka \\
  University of Southern California \\
  Los Angeles, CA, USA \\
  \texttt{nagaharshitamarupaka@gmail.com} \\}

\begin{document}
\maketitle
\begin{abstract}

Technical reports and articles often contain valuable information in the form of semi-structured data like charts, and figures. Interpreting these and using the information from them is essential for downstream tasks such as question answering (QA). Current approaches to visual question answering often struggle with the precision required for scientific data interpretation, particularly in handling numerical values, multi-step reasoning over visual elements, and maintaining consistency between visual observation and textual reasoning. We present our approach to the SciVQA 2025 shared task, focusing on answering visual and non-visual questions grounded in scientific figures from scholarly articles. 

% Our methodology combines prompt engineering, and ensemble modeling of multiple multimodal small language models (MSLMs). Through error analysis on validation split, we demonstrate that our ensemble approach achieves significant improvements over individual models. Our best-performing system achieved a ROUGE-1 and ROUGE-L F1 score of \textbf{0.740} on the SciVQA test split.

% TODO: to be reviewed, add the BertScore 
We conducted a series of experiments using models with 5B to 8B parameters. Our strongest individual model, InternVL3, achieved ROUGE-1 and ROUGE-L F1 scores of \textbf{0.740} and a BERTScore of \textbf{0.983} on the SciVQA test split. We also developed an ensemble model with multiple vision language models (VLMs). Through error analysis on the validation split, our ensemble approach improved performance compared to most individual models, though InternVL3 remained the strongest standalone performer. Our findings underscore the effectiveness of prompt optimization, chain-of-thought reasoning and ensemble modeling in improving the model's ability in visual question answering.
\end{abstract}

\section{Introduction}

Scientific literature communicates complex ideas not only through text but also through carefully designed visual elements including charts, graphs, diagrams, and technical illustrations. These visualizations serve as dense information carriers, encoding quantitative relationships, experimental results, architectural designs, and conceptual frameworks that are essential for scientific understanding. The ability to automatically interpret and reason about these visual elements represents a critical challenge in advancing scientific AI systems.

The task of Visual Question Answering (VQA) over scientific figures presents unique challenges that distinguish it from general-domain VQA. Scientific visualizations demand mathematical precision, often requiring exact numerical extraction and calculation. They involve complex compositional reasoning across multiple visual elements, and frequently contain domain-specific conventions, symbols, and representations that require specialized understanding \cite{Ishmam_2024}. Furthermore, scientific figures often embed multiple layers of information, including raw data points, derived trends, statistical relationships, and comparative analyses.

Current VQA models, while showing impressive performance on general datasets, often struggle with the precision and reasoning depth required for scientific applications \cite{kabir2024comprehensivesurveyvisualquestion}. Common failure modes include visual grounding errors, where models misinterpret chart elements or scales; compositional reasoning failures, where multi-step logical processes break down; and consistency issues between visual observations and textual explanations \cite{tanjim2025exploringrewritingapproachesdifferent, thawakar2025llamavo1rethinkingstepbystepvisual}.

This paper presents our approach to the SciVQA Shared Task\footnote{\url{https://sdproc.org/2025/scivqa.html}} \footnote{
\url{https://huggingface.co/datasets/katebor/SciVQA}
} \cite{borisova-scivqa-2025}, focusing on QA over scientific visualizations. The task involves answering closed-ended visual (i.e., addressing visual attributes such as colour, shape, size, height, etc.) and non-visual (not addressing figure visual attributes) questions. We leverage the reasoning and visual understanding capabilities of VLMs, and employ task-specific Chain-of-Thought (CoT) \cite{wei2023chainofthoughtpromptingelicitsreasoning} prompting techniques to retrieve and summarize relevant information from the visualizations. Our approach involved testing multiple prompt variants and selecting optimal configurations based on validation performance. To support reproducibility and future research, we make the code publicly available on GitHub\footnote{\url{https://github.com/NagaHarshita/Infyn-SciVQA}
}.

% \textcolor{blue}{We did strategic experiments spanning around prompt engineering (with and without CoT), Ensemble modelling and majority voting to obtain the best results} 

% \textcolor{red}{Our approach demonstrates that a single high-performance multimodal model with advanced vision encoding capabilities, when combined with systematic prompt engineering, can achieve state-of-the-art results without the complexity of ensemble methods.}
Our main contributions are:
\begin{itemize}[itemsep=2pt, parsep=0pt, topsep=2pt]
    \item A systematic ensemble strategy with figure type specific model selection based on comprehensive validation analysis.
    \item Optimized prompt engineering templates tailored to different question answer pair and figure type combinations.
\end{itemize}

\section{Related Work}

Recent advancements in chart-based QA have focused on various approaches to understanding and generating responses about visualizations. ChartLlama \cite{han2023chartllamamultimodalllmchart} and UniChart \cite{masry2023unichartuniversalvisionlanguagepretrained} demonstrate the benefits of chart-specialized language models, showing improved performance in both chart captioning and QA tasks. These works often rely on explicit chart structure parsing as a preprocessing step, achieving strong results on synthetic chart datasets.

% \cite{zheng2024advancingchartquestionanswering} highlighted the importance of accurate chart component recognition to boost QA performance in chart-based VQA, emphasizing explicit parsing as a key preprocessing step. In contrast, our method focuses on general-purpose vision-language models adapted via sophisticated reward modeling without requiring explicit structural parsing.

% ChartVLM \cite{xia2025chartxchartvlmversatile} introduces a dual-stream vision-language model specifically designed for complex chart understanding. It combines structural parsing with multimodal reasoning to handle multi-type charts and diverse question formats, showcasing strong performance across both synthetic and real-world benchmarks, including figure-based QA in scientific literature.

% Matcha-Chart2text-pew \cite{liu2022matcha}, an extension of Google's Matcha model, leverages encoder-decoder architectures pre-trained on large-scale chart datasets for chart-to-text generation. It has demonstrated impressive capabilities in summarizing chart semantics into coherent textual descriptions, offering insights applicable to both caption generation and answer synthesis in VQA pipelines.

Chart-based Reasoning \cite{carbune2024chartbasedreasoningtransferringcapabilities} and LlamaV-o1 \cite{thawakar2025llamavo1rethinkingstepbystepvisual} propose decomposed reasoning traces and transfer of LLM capabilities to visual settings. Our work builds on these insights by emphasizing structured reasoning and adopt step-level supervision to encourage coherent and faithful intermediate reasoning in visual contexts.

Other relevant works include SPIQA \cite{pramanick2025spiqadatasetmultimodalquestion} and MathVista \cite{lu2024mathvistaevaluatingmathematicalreasoning}, which evaluate visual reasoning in scientific domains. MathVista particularly focuses on precise numerical and symbolic interpretation in mathematical visualizations, similar to our emphasis on scientific accuracy.

% Recent work on visual hallucination mitigation, prompt-driven harmonization, and geometry-aware prompting strategies \cite{Liu_Zhu_Wen_Lu_Lin_Chen_2025} inform our approach to harmonizing visual and textual modalities for more grounded answers. We extend these concepts through our multi-dimensional reward system that explicitly evaluates visual engagement and grounding.

In the realm of prompt engineering and model alignment, \cite{zhan2025sprialigninglargelanguage} proposed SPRI (Situated-PRInciples), a framework that automatically generates context-specific guiding principles for each input query to improve model alignment. Their approach demonstrates that instance-specific principles can outperform generic ones, which informs our ensemble methodology that combines prompt engineering with multiple VLMs.

Motivated by recent advances in prompt rewriting \cite{tanjim2025exploringrewritingapproachesdifferent}, we explore instruction tuning and prompt optimization to enhance model adherence to scientific QA formats. Unlike previous work that focuses on architectural innovations requiring additional training, our approach focuses on ensemble strategies and prompt optimization for maximum performance on scientific VQA tasks.

% Motivated by recent advances in prompt rewriting \cite{tanjim2025exploringrewritingapproachesdifferent}, we explore instruction tuning to enhance model adherence to scientific QA formats.

\section{Dataset}

% The SciVQA dataset comprises of scientific figures from ACL Anthology and arXiv papers. Each figure is annotated with seven QA pairs and includes metadata such as caption, figure ID, figure type (e.g., compound, line graph, bar chart, scatter plot), QA pair type. More details about the dataset distribution can be seen in Section \ref{sec:appendix-dataset-distribution}, and Tables \ref{table:1}, \ref{table:2}, and \ref{table:3} in the Appendix.

The SciVQA dataset comprises scientific figures from ACL Anthology and arXiv papers. Each figure is annotated with seven question-answer pairs and associated metadata including captions, figure IDs, figure types (e.g., compound, line graph, bar chart, scatter plot), and QA pair types, with dataset splits and distributions detailed in Section~\ref{sec:appendix-dataset-distribution} and Tables~\ref{table:1}, \ref{table:2}, and \ref{table:3}.

\section{Methodology}

Our system integrates three key components:

\begin{itemize}[itemsep=2pt, parsep=0pt, topsep=2pt]
    \item systematic prompt optimization for different figure types
    \item strategic ensemble modeling, and
    \item post-processing for answer standardization.
\end{itemize}

We utilized the vLLM \cite{kwon2023efficientmemorymanagementlarge} engine for maximum compute utilization during inference. A40 instances were sufficient for 7B models, while 8B models required A100 GPUs. CoT inference required approximately twice the computation time due to the two-level reasoning process, but provided significant quality improvements.

\subsection{Model Selection}
To inform model selection and ensure alignment with the target domain, we referred to the performance of recent models on established multimodal QA benchmarks analogous to SciVQA \cite{borisova-scivqa-2025}, including ChartQA \cite{masry2022chartqa}, MathVista \cite{lu2024mathvistaevaluatingmathematicalreasoning}, ChartXiv \cite{wang2024charxiv}. 
VLMs in the 5–8B parameter range demonstrated competitive performance on these leaderboards, achieving results comparable to significantly larger models with 32–72B parameters.

According to the InternVL3 technical report \cite{zhu2025internvl3exploringadvancedtraining}, the models InternVL3-8B and Qwen2.5-VL-7B performed well on tasks such as OCR, chart, and document understanding, specifically on datasets like ChartXiv and ChartQA. Additionally, a fine-tuned version of the Qwen2.5-VL-7B Instruct model \cite{bai2025qwen25vltechnicalreport}, as reported in the Bespoke technical report \footnote{\url{https://www.bespokelabs.ai/blog/bespoke-minichart-7b}}, demonstrates competitive performance on ChartXiv, ChartQA, and EvoChart, achieving results comparable to InternVL3-8B. Based on these observations, we chose the following four VLMs, namely, InternVL3-8B, Qwen2.5-VL-7B Instruct, Bespoke MiniChart 7B, and Phi-4 Multimodal Instruct for our task.

\noindent \textbf{InternVL3-8B} \cite{zhu2025internvl3exploringadvancedtraining} features an advanced vision encoder architecture tailored for complex visual understanding, unlike Qwen2.5-VL which uses a standard ViT encoder \cite{zhu2025internvl3exploringadvancedtraining}. The model supports high-resolution image processing capabilities essential for interpreting detailed charts, and incorporates multi-scale feature extraction to enable both global comprehension and fine-grained numerical reading. It is particularly robust when handling overlapping text and visually dense layouts commonly found in scientific figures. \textbf{Qwen2.5-VL-7B Instruct} \cite{bai2025qwen25vltechnicalreport} exhibits strong mathematical reasoning, although its performance is limited by the underlying vision encoder. \textbf{Bespoke MiniChart 7B}, trained with DPO \cite{rafailov2024directpreferenceoptimizationlanguage}, benefits from improved chain-of-thought reasoning for chart understanding tasks, but lacks architectural features suited for complex scientific visualizations. Finally, \textbf{Phi-4 Multimodal Instruct} (5.6B) \cite{microsoft2025phi4minitechnicalreportcompact} offers general multimodal capabilities but is not specifically optimized for scientific content.

\subsection{Prompt Optimization}
We crafted task-specific prompts that incorporate captions, figure types, and QA pair types. Prompt variants included explicit CoT cues, multiple correct answer hints, and image-caption-context fusion, with performance differences noted across QA pair types. Additionally, we set two baseline models (both using InternVL3): one using a general prompt without specifying the expected output format, and another with explicit formatting instructions stating that the output should be either a number or a single sentence. Baseline 1 refers to a general prompt without formatting constraints, while Baseline 2 uses explicit answer formatting instructions (exact prompts provided in the Appendix in Table \ref{tab:prompt-table-baseline}). The structured format ensures consistency and enables automated evaluation of both reasoning quality and final answers. All the components described below (e.g., Base Prompt, Compound Images Prompt, Figure Type Prompt, etc.), and in the Tables \ref{tab:prompt-table-single} and \ref{tab:prompt-table-cot} are combined into a single, composite prompt to ensure that all possible aspects of the task are considered.

\subsubsection{Single Prompt}
% We developed an initial prompt that includes the figure caption, question answer pair type classification and task-specific instructions that elicit reasoning.  The exact prompt for each of the above are in Table  \ref{tab:prompt-table-single}.

We developed an initial prompt that includes the figure caption, question-answer pair type classification, and task-specific instructions that elicit reasoning, with exact prompts detailed in Table~\ref{tab:prompt-table-single}.

\noindent \textbf{Prompt Used:} \textit{Base Prompt + Compound Images Prompt + Figure Type Prompt + Question + Binary Prompt + Choice Prompt}

\subsubsection{CoT and Rethink}
% Incorporating Chain-of-Thought (CoT) \cite{wei2023chainofthoughtpromptingelicitsreasoning} and Rethink mechanisms \cite{wang2025vlrethinkerincentivizingselfreflectionvisionlanguage} where models regenerate answers with self-correction, significantly enhances performance, particularly for math-intensive and ambiguous examples. Prompts are designed to elicit reflective thinking, with final answers distinctly highlighted using structured XML tags ($<reasoning>$ and $<answer>$). The exact prompt for each of the above are in Table \ref{tab:prompt-table-cot}.

Incorporating Chain-of-Thought (CoT) \cite{wei2023chainofthoughtpromptingelicitsreasoning} and Rethink mechanisms \cite{wang2025vlrethinkerincentivizingselfreflectionvisionlanguage} where models regenerate answers with self-correction significantly enhances performance, particularly for math-intensive and ambiguous examples. Prompts are designed to elicit reflective thinking, with final answers distinctly highlighted using structured XML tags ($<reasoning>$ and $<answer>$), as detailed in Table~\ref{tab:prompt-table-cot}.

\noindent \textbf{Step 1 Prompt Used:} \textit{Step 1 Base Prompt + Compound Images Prompt}

\noindent \textbf{Step 2 Prompt Used:} \textit{Step 2 Base Prompt + Figure Type Prompt + Binary Prompt + Choice Prompt}

\subsection{Ensemble}
Based on a comprehensive validation analysis (see Section \ref{sec:appendix-error-analysis} and Table \ref{tab:ensemble-metrics-1} in Appendix for detailed model performance across figure types), we implemented a figure-type-aware ensemble approach in which each model was assigned to chart types aligned with its demonstrated strengths. Specifically, Qwen2.5-VL was selected for scatter plots, confusion matrices, trees, and graphs, given its relative effectiveness on relational and structural visualizations. Bespoke MiniChart was applied to pie charts, bar charts, architecture diagrams, neural networks, and box plots, leveraging its finetuning for specialized chart comprehension. Meanwhile, Phi-4 was assigned to line charts, tables, histograms, vector plots, and illustrative diagrams, where it showed comparatively better performance. Although this targeted ensemble method yielded competitive results, it was ultimately outperformed by InternVL3, which demonstrated robust and consistent accuracy across all figure types.

\subsection{Postprocessing}
Our post-processing pipeline involved two key modifications to improve answer quality and evaluation metrics. First, all $\abs{end}$ tags were removed from generated responses to ensure clean output format. Then, for questions where the reasoning process determined insufficient information to give a valid response, outputs were standardized to "It is not possible to answer this question based only on the provided data." regardless of the initial model output. Model outputs, after applying post-processing, were evaluated on both the test set (Table~\ref{tab:wide-metrics-2}) and validation set (Table~\ref{tab:wide-metrics-1}) using BERTScore and ROUGE metrics.

\begin{table*}[ht]
\centering
\renewcommand{\arraystretch}{1.0}
\begin{tabularx}{\textwidth}{p{2.2cm}XXXXXXXXX}
\toprule
\textbf{Model} & \textbf{\mbox{R1-F1}} & \textbf{\mbox{R1-P}} & \textbf{\mbox{R1-R}} & \textbf{\mbox{RL-F1}} & \textbf{\mbox{RL-P}} & \textbf{\mbox{RL-R}} & \textbf{\mbox{BS-F1}} & \textbf{\mbox{BS-P}} & \textbf{\mbox{BS-R}} \\
\midrule
InternVL3 & \textbf{0.730} & \textbf{0.744} & \textbf{0.732} & \textbf{0.729} & \textbf{0.743} & \textbf{0.731} & \textbf{0.981} & \textbf{0.983} & \textbf{0.980} \\
Qwen2.5-VL & 0.619 & 0.621 & 0.641 & 0.618 & 0.620 & 0.641 & 0.970 & 0.967 & 0.973 \\
Bespoke & 0.636 & 0.641 & 0.647 & 0.634 & 0.640 & 0.645 & 0.975 & 0.975 & 0.976 \\
% LlamaV-o1 & & & & & & & & & \\
Phi-4 & 0.532 & 0.531 & 0.596 & 0.531 & 0.529 & 0.595 & 0.950 & 0.944 & 0.956 \\
Ensemble & \underline{0.646} & \underline{0.651} & \underline{0.660}  & \underline{0.645} & \underline{0.650} & \underline{0.658} & \underline{0.974} & \underline{0.974} & \underline{0.976} \\
\bottomrule
\end{tabularx}
\caption{Comparison across ROUGE (R1, RL) and BERTScore (BS) metrics (F1, Precision, Recall) on \textbf{validation (without CoT)} after applying post-processing.}
\label{tab:wide-metrics-1}
\end{table*}

% \FloatBarrier
\begin{table*}
\centering
\renewcommand{\arraystretch}{1.0}
\begin{tabularx}{\textwidth}{p{2.2cm}XXXXXXXXX}
\toprule
\textbf{Model} & \textbf{R1-F1} & \textbf{R1-P} & \textbf{R1-R} & \textbf{\makebox[2.2em][c]{RL-F1}} & \textbf{RL-P} & \textbf{RL-R} & \textbf{BS-F1} & \textbf{BS-P} & \textbf{BS-R} \\
\midrule
InternVL3 & \textbf{0.740} & \textbf{0.754} & \textbf{0.739} & \textbf{0.740} & \textbf{0.754} & \textbf{0.738} & \textbf{0.983} & \textbf{0.985} & \textbf{0.982} \\
Qwen2.5-VL & 0.695 & 0.699 & 0.714 & 0.694 & 0.698 & 0.713 & 0.975 & 0.973 & 0.977 \\
Bespoke & 0.709 & 0.716 & 0.716 & 0.708 & 0.715 & 0.715 & 0.979 & 0.979 & 0.979 \\
Phi-4 & 0.562 & 0.566 & 0.578 & 0.561 & 0.565 & 0.578 & 0.969 & 0.966 & 0.970 \\
Ensemble & \underline{0.735} & \underline{0.744} & \underline{0.744} & \underline{0.734} & \underline{0.743} & \underline{0.743} & \underline{0.979} & \underline{0.978} & \underline{0.980} \\
\bottomrule
\addlinespace[0.65ex]
InternVL3 & 0.727 & 0.739 & 0.728 & 0.727 & 0.738 & 0.727 & \underline{0.982} & \underline{0.983} & \underline{0.981} \\
Qwen2.5-VL & 0.633 & 0.633 & 0.658 & 0.633 & 0.632 & 0.658 & 0.972 & 0.969 & 0.975 \\
Bespoke & 0.652 & 0.657 & 0.664 & 0.651 & 0.656 & 0.663 & 0.976 & 0.976 & 0.977 \\
Phi-4 & 0.544 & 0.540 & 0.600 & 0.543 & 0.540 & 0.600 & 0.954 & 0.948 & 0.960 \\
Ensemble & 0.705 & 0.714 & 0.710 & 0.704 & 0.713 & 0.709 & 0.979 & 0.979 & 0.979 \\
Baseline 1 & 0.180 & 0.164 & 0.498 & 0.180 & 0.163 & 0.496 & 0.834 & 0.812 & 0.857 \\
Baseline 2 & 0.700 & 0.707 & 0.710 & 0.699 & 0.707 & 0.710 & 0.977 & 0.977 & 0.978 \\
\bottomrule
\end{tabularx}
\caption{Comparison across ROUGE (R1, RL) and BERTScore (BS) metrics (F1, Precision, Recall) on \textbf{test with CoT (top) and without CoT (bottom)} after applying post-processing.}
\label{tab:wide-metrics-2}
\end{table*}
% \FloatBarrier

\subsection{Results}
% Validation results provided crucial insights for both prompt fine-tuning and reward function optimization, enabling iterative improvements to the overall system. 

% CoT prompting consistently improves performance across all models, with particularly strong gains on complex multi-step reasoning questions (average improvement of X\% in ROUGE-1 F1).

% Between with or without CoT across all MSLMs, CoT and rethink performed consistenctly well. Another obvious observation is that more parameter model gives better reasoning and answers. 
% Best model performed better than other models by atleast +0.2 Rouge-1 F1. Other individual models when used our ensemble technique gave better Rouge-1 F1 by atleast 0.15 score. 

% The structured reasoning approach led to:

% \begin{itemize}[itemsep=2pt, parsep=0pt, topsep=2pt]
%     \item Improved accuracy on complex multi-step problems
%     \item More consistent format adherence
%     \item Better visual grounding in responses
%     \item Enhanced ability to self-correct initial misconceptions
% \end{itemize}

Our final system, based on an ensemble approach, was submitted to the challenge leaderboard and ranked 5th. Table \ref{tab:leaderboard} in Appendix shows the top-7 rankings as on the leaderboard.

\noindent \textbf{Chain-of-Thought Performance:} CoT prompting achieved consistent improvements across all VLMs on the test set, with gain in scores for complex multi-step reasoning questions. CoT with rethinking mechanisms demonstrated the most stable performance across different question types.

\noindent \textbf{Model Scale Impact:} Larger parameter models consistently outperformed smaller variants on the test set, confirming the correlation between model capacity and reasoning quality in scientific visual question answering.

\noindent \textbf{Comparative Performance:} InternVL3 achieved the highest individual model performance, outperforming other individual models by at least +0.30 ROUGE-1 F1 score on the test split. 

% Our ensemble methodology improved individual model performance by at least +0.15 ROUGE-1 F1 score.

% Structured Reasoning Effectiveness: Test results confirmed that structured reasoning approaches delivered:
% \begin{itemize}[itemsep=2pt, parsep=0pt, topsep=2pt]
% \item Enhanced accuracy on complex multi-step scientific problems
% \item Improved consistency in answer format adherence
% \item Better visual grounding in chart interpretation responses
% \item Superior error correction capabilities during reasoning
% \end{itemize}

% \textcolor{red}{scoring.py inconsistencies with validation results ['A', 'B'], five}

% \textcolor{red}{add some examples/samples from the dataset and the model outputs}

% \textcolor{red}{table showing percentage improvement in performance with and without CoT}

\section{Conclusion and Future Work}
This work demonstrates that advanced vision encoding architecture, combined with systematic prompt engineering, provides a highly effective approach to scientific visual question answering. High-quality visual understanding is critical for strong performance. Our approach establishes a strong baseline for the SciVQA dataset with a ROUGE-1 and ROUGE-L F1 score of 0.740.
% Rather than pursuing complex ensemble methods or reasoning frameworks, focusing on advanced vision architectures and task-specific optimization yields better results.
% Our analysis shows that InternVL3's superior vision encoder capabilities enable it to outperform more complex ensemble approaches while maintaining computational efficiency.

% The key insight from our work is that for scientific VQA tasks, the quality of visual understanding fundamentally determines performance. Rather than pursuing complex ensemble methods or reasoning frameworks, focusing on advanced vision architectures and task-specific optimization yields better results.

\subsection{Future Directions}

% \noindent \textbf{Using LLM-as-a-Judge}
% For samples with low confidence scores, we plan to implement feedback loops using LLM-as-a-Judge \cite{zheng2023judgingllmasajudgemtbenchchatbot} approaches. Confidence scoring will be based on consistency across multiple inference runs, with samples showing answer variation tagged for additional review.

% \noindent \textbf{Training data augmentation}
% Creating synthetic chart-question pairs and incorporating diverse datasets with various chart and figure types will enhance performance on low-frequency question types. Currently, reasoning traces are generated only for challenging validation samples ($\sim$300 examples), and extending this to training data should yield further improvements.

\noindent \textbf{Advanced Reasoning Techniques:}
Exploring advanced prompting techniques such as Tree-of-Thought \cite{yao2023treethoughtsdeliberateproblem}, leveraging Mixture-of-Experts (MoE) \cite{shazeer2017outrageouslylargeneuralnetworks} models tailored to different question-answer pairs, and incorporating expert-critic ensembles for answer re-ranking holds significant potential for enhancing overall performance.

\noindent \textbf{Scalability and Model Improvements:}
Running inference using larger models or fine-tuning the current models on this dataset, enabled by increased computational resources, is expected to yield substantial performance gains.

\noindent \textbf{Quality of the Dataset:}
Addressing the identified dataset quality issues from Appendix \ref{sec:appendix-data-quality} by standardizing data formatting (e.g., multi-correct answers, numerical representations, and answer formatting) and conducting a comprehensive review of the gold standard annotations, to ensure more accurate evaluations. These improvements will enhance dataset accuracy, provide fairer evaluations, and help ensure more reliable model performance comparisons in subsequent iterations of this task.

% \noindent \textbf{Code Repository:} Our complete implementation including the GRPO reward functions, prompt templates, fine-tuning scripts, and evaluation pipelines is available at \textcolor{red}{URL here}.

% \noindent \textbf{Model Weights:} Fine-tuned model checkpoints for Bespoke MiniChart 7B and InternVL3 8B (both SFT and GRPO variants) are released through HuggingFace Model Hub. These models can serve as strong baselines for future work in scientific VQA.

% \noindent \textbf{Reasoning Traces Dataset:} We release the 300 challenging validation examples with high-quality reasoning traces generated by GPT-4o and Claude 3.7 Sonnet. This dataset can be valuable for training other models or developing alternative reward functions.

\section*{Limitations}
Due to computational resource constraints, experiments were primarily limited to 7B model variants. Larger model variants and fine-tuning using the entire train split would most definitely yield superior results.

\bibliography{custom}
\bibliographystyle{acl_natbib}
\clearpage
\newpage
\appendix
\section*{Appendix}
\section{Dataset Distribution}
\label{sec:appendix-dataset-distribution}

The dataset is biased toward line charts (66\%), requiring strong numerical reading capabilities. Additionally, a high percentage of non-visual questions (around 60\%) highlights the need for reasoning that goes beyond visual features.
% Finally, the presence of unanswerable questions (14.3\%) indicates the importance of enabling models to recognize when information is insufficient and to avoid hallucinations.

% \vspace{0.2cm}

\begin{center}
\begin{tabular}{|l|c|}
\hline
\textbf{Dataset Split} & \textbf{Samples} \\ \hline
Train         & 15,120 \\ \hline
Validation    & 1,680 \\ \hline
Test          & 4,200 \\ \hline
\end{tabular}
\captionof{table}{Dataset split with number of samples.}
\label{table:1}
\end{center}

% \vspace{-1cm}

\begin{center}
\renewcommand{\arraystretch}{1.2}
\begin{tabularx}{\columnwidth}{|c|c|X|c|}
\hline
\textbf{QA Type} & \multicolumn{2}{c|}{\textbf{Answer Set}} & \textbf{Samples} \\
\hline
\multirow{6}{*}{\shortstack{Closed\\-ended}} 
  & \multirow{2}{*}{Infinite} 
    & Visual & 1,079 \\ \cline{3-4}
  &  
    & Non-visual & 2,172 \\ \cline{2-4}
  & \multirow{4}{*}{Finite} 
    & Binary \& Visual & 1,124 \\ \cline{3-4}
  &  
    & Binary \& Non-visual & 3,219 \\ \cline{3-4}
  &  
    & Non-binary \& Visual & 1,751 \\ \cline{3-4}
  &  
    & Non-binary \& Non-visual & 3,615 \\ \hline
\multicolumn{3}{|c|}{Unanswerable} & 2,160 \\ \hline
\end{tabularx}
\captionof{table}{QA Pair Type Categorization in the Train split.}
\label{table:2}
\end{center}

% \vspace{0.5cm}

\begin{center}
\begin{tabular}{|l|c|}
\hline
\textbf{Figure Type} & \textbf{Samples} \\ \hline
Line Chart         & 10,007 \\ \hline
Tree         & 924 \\ \hline
Scatter Plot    & 735 \\ \hline
Graph         & 553 \\ \hline
Bar Chart          & 525 \\ \hline
Architecture Diagram          & 504 \\ \hline
Pie Chart    & 497 \\ \hline
Neural Networks    & 462 \\ \hline
Confusion Matrix          & 427 \\ \hline
Box Plot         & 133 \\ \hline
Histogram    & 77 \\ \hline
Other          & 77 \\ \hline
\end{tabular}
\captionof{table}{Figure Type Distribution in the Train split.}
\label{table:3}
\end{center}

\section{Error Analysis}
\label{sec:appendix-error-analysis}
We conducted error analysis manually on the validation dataset because gold answers are available for it. We only selected incorrect predictions and categorized them into three primary types:

% would be nice to add the percentage of each type of error as well

\noindent \textbf{Visual Misinterpretations:} Issues with feature extraction from images, including:

\begin{itemize}[itemsep=2pt, parsep=0pt, topsep=2pt]
    \item Comparing sub-figures on different scales
    \item Misunderstanding axis starting points
    \item Difficulties with overlapping text
    \item Challenges with low-resolution images
\end{itemize}

\noindent \textbf{Numerical Misalignments:} Precision issues in numerical extraction and calculation, often stemming from visual ambiguity in chart elements.

\noindent \textbf{Flawed Reasoning:} Instances where logical progression was incorrect despite proper visual observation, or cases where correct reasoning led to incorrect answers due to misalignment with reference answer formats.

Notably, some failures occurred in compound charts and arose from misinterpreted visual cues or insufficient numerical precision. The use of more powerful vision encoders could address many of these issues and further improve performance.

\section{Dataset Quality Issues}
\label{sec:appendix-data-quality}

During our validation analysis, we identified several systematic inconsistencies in the gold standard annotations that may impact evaluation reliability:

\noindent \textbf{Format Inconsistencies:}
\begin{itemize}[itemsep=2pt, parsep=0pt, topsep=2pt]
    \item Multi-correct answers appear in inconsistent formats: ["A", "B"] instead of the expected A,B format
    \item Numerical representations vary between word form ("three") and digit form ("3") within similar contexts
    \item Answer formatting lacks standardization across question types
\end{itemize}

\noindent \textbf{Annotation Errors:} We identified potential annotation errors through manual inspection. For example: instance\_id 09dab5a715034cebb2a62f0f1c2a75c9 gold answer is "52,3\%" but visual inspection suggests that the correct answer should be "3\%" or "3", which our models corrects predict.

These inconsistencies suggest that reported performance metrics may underestimate true model capabilities, as models may be penalized for providing correct answers that don't match inconsistent gold standards. A comprehensive gold standard review and standardization would benefit future iterations of this shared task.

\section{Alternative Approaches Evaluated}
\label{sec:appendix-alternative-approaches}

% \noindent \textbf{Ensemble Methods:} We initially explored ensemble approaches combining multiple models but found that InternVL3's individual performance exceeded ensemble results while maintaining lower computational overhead.

\noindent \textbf{Majority Voting:} Simple majority voting across models showed minimal improvement over InternVL3 alone, confirming the quality-over-quantity principle.

\noindent \textbf{Fine-tuning:} Limited computational resources prevented extensive fine-tuning, but initial experiments suggested that InternVL3's pre-trained capabilities were already well-suited for the task. To enhance performance, we performed supervised fine-tuning (SFT) using LoRA \cite{hu2021loralowrankadaptationlarge}, targeting all linear layers and the vision encoder, on a subset of 5,000 training samples for the Bespoke MiniChart model, which yielded a slight performance improvement. We are currently extending this approach by applying Group Relative Policy Optimization (GRPO)-based \cite{deepseek-math} fine-tuning in a similar fashion.

\section{Tables}
\label{sec:appendix-tables}

\begin{table}[H]
\centering
\begin{tabular}{|l|p{5cm}|}
\hline
\textbf{Baseline} & \textbf{Prompt Content} \\ \hline
\textbf{1} & 
You are a helpful assistant. Give the concise answer for the context given below. The caption of the figure is mentioned as, [caption]. The question for the figure is, [question]\\
\hline
\textbf{2} & 
Answer the question with only the raw numerical value or single word/phrase, omitting all units, context words, and explanatory text. The caption of the figure is mentioned as, [caption]. The question for the figure is, [question]\\
\hline
\end{tabular}
\caption{Baseline Prompts}
\label{tab:prompt-table-baseline}
\end{table}

\begin{table*}[t]
\centering
\renewcommand{\arraystretch}{1.2}
\begin{tabular}{|p{2.2cm}|p{13cm}|}
\hline
\textbf{Prompt Type} & \textbf{Prompt Content} \\
\hline
\textbf{Base} & 
Answer the question with only the raw numerical value or single word/phrase, omitting all units, context words, and explanatory text. The caption of the figure is mentioned as, [caption]. \\
\hline
\textbf{Compound Images} & 
This is a compound figure containing multiple subfigures. Navigate to [fig\_numb] graph in the compound figure to answer the question. \\
\hline
\textbf{Figure Type} & 
\textbf{Line Chart:} 

Focus on the following aspects of the line chart: 
\begin{itemize}[itemsep=2pt, parsep=0pt, topsep=2pt]
    \item Colors of different lines and their meanings
    \item X and Y axis labels and their units
    \item Scale and range of values
    \item Trends and patterns in the lines
\end{itemize}
\textbf{Bar Chart:}

Focus on the following aspects of the bar chart:
\begin{itemize}[itemsep=2pt, parsep=0pt, topsep=2pt]
    \item Colors of different bars and their meanings
    \item X and Y axis labels and their units
    \item Scale and range of values
    \item Height and position of bars
\end{itemize}
\textbf{Box Plot:}

Focus on the following aspects of the box plot:
\begin{itemize}[itemsep=2pt, parsep=0pt, topsep=2pt]
    \item Median line position
    \item Box boundaries (Q1 and Q3)
    \item Whisker extent
    \item Outliers if present
\end{itemize}
\textbf{Confusion Matrix:}

Focus on the following aspects of the confusion matrix:
\begin{itemize}[itemsep=2pt, parsep=0pt, topsep=2pt]
    \item Row and column labels
    \item Numerical values in each cell
    \item Color intensity if present
    \item Overall distribution of values
\end{itemize}
\textbf{Pie Chart:}

Focus on the following aspects of the pie chart:
\begin{itemize}[itemsep=2pt, parsep=0pt, topsep=2pt]
    \item Segments and their labels
    \item Percentage or proportion values
    \item Colors of different segments
    \item Size of each segment relative to others
\end{itemize}
\textbf{Others:}

Focus on the following aspects of the figure:
\begin{itemize}[itemsep=2pt, parsep=0pt, topsep=2pt]
    \item Colors and the labels present in the figure
    \item Any other relevant information present in the figure
\end{itemize} \\
\hline
\textbf{Binary} & 
This is a binary question. Answer with ‘Yes’ or ‘No’ based on [visual/textual] evidence. Respond affirmatively only if supported. \\
\hline
\textbf{Choice} & 
Return only the corresponding letter(s) of the correct answer(s). Only output the letter(s) corresponding to the correct choice. [answer\_choices] \\
\hline
\end{tabular}
\caption{Instruction Prompts for Single Prompt}
\label{tab:prompt-table-single}
\end{table*}

\begin{table*}[t]
\centering
\renewcommand{\arraystretch}{1.2}
\begin{tabular}{|p{2.5cm}|p{12cm}|}
\hline
\textbf{Prompt Type} & \textbf{Prompt Content} \\
\hline
\textbf{Step 1 Base Prompt} & 
STEP 1: INITIAL ANALYSIS

Given the figure, caption, and question, analyze and answer step by step.
Regularly perform self-questioning, self-verification, self-correction to check your ongoing reasoning, using connectives such as "Wait a moment", "Wait, does it seem right?" etc.

Caption: [caption]

Question: [question]

Analyse the key visual elements (lines, shapes, colors) that address the question and analyze the relationships between elements. Then, extract the specific numerical/positional information from the figure and caption to answer the question.
\\
\hline
\textbf{Compound Images Prompt} & 
Same as single prompt \\
\hline
\textbf{Step 2 Base Prompt} & 

STEP 2: COT INFERENCE

Answer the question with only the raw numerical value or single word/phrase, omitting all units, context words, and explanatory text. Approximations in the scale are allowed. \\
\hline
\textbf{Figure Type Prompt} & 
Same as single prompt \\
\hline
\textbf{Binary Prompt} & 
Same as single prompt \\
\hline
\textbf{Choice Prompt} & 
\textit{(For non-binary finite answer sets)}: Based on the reasoning above, match it to one or more of the provided answer options: [answer\_choices]

Return only the corresponding letter(s) of the correct answer(s). 
Do not explain your choice, do not rephrase the answer, and do not repeat the option text. 
Only output the letter(s) corresponding to the correct choice. 
If multiple letters are correct, separate them by commas without spaces (for example: B,C). 
If all options are correct, return A,B,C,D. 
Do not add anything else. \\
\hline
\end{tabular}
\caption{Instruction Prompts for CoT}
\label{tab:prompt-table-cot}
\end{table*}

\begin{table*}[ht]
\centering
\renewcommand{\arraystretch}{1.0}
\begin{tabularx}{\textwidth}{p{3.5cm} *{8}{>{\centering\arraybackslash}X}}
\toprule
\textbf{Chart Type} & \multicolumn{2}{c}{\textbf{Bespoke}} & \multicolumn{2}{c}{\textbf{InternVL3}} & \multicolumn{2}{c}{\textbf{Qwen2.5-VL}} & \multicolumn{2}{c}{\textbf{Phi-4}} \\
& Acc. (mean) & Std. Dev. & Acc. (mean) & Std. Dev. & Acc. (mean) & Std. Dev. & Acc. (mean) & Std. Dev. \\
\midrule
line\_chart               & 54.23 & 49.82 & 63.97 & 48.01 & 50.68 & 50.00 & 42.40 & 49.42 \\
line\_chart,table         & 42.86 & 49.49 & 85.71 & 34.99 & 42.86 & 49.49 & 57.14 & 49.49 \\
tree                      & 56.19 & 49.62 & 61.90 & 48.56 & 53.33 & 49.89 & 44.76 & 49.72 \\
scatter\_plot             & 55.71 & 49.67 & 70.00 & 45.83 & 57.14 & 49.49 & 40.00 & 48.99 \\
pie\_chart                & 67.35 & 46.89 & 73.47 & 44.15 & 67.35 & 46.89 & 44.90 & 49.74 \\
architecture\_diagram     & 67.86 & 46.70 & 76.79 & 42.22 & 55.36 & 49.71 & 28.57 & 45.18 \\
box\_plot                 & 50.00 & 50.00 & 50.00 & 50.00 & 50.00 & 50.00 & 35.71 & 47.92 \\
neural\_networks          & 62.50 & 48.41 & 71.43 & 45.18 & 58.93 & 49.20 & 32.14 & 46.70 \\
confusion\_matrix         & 54.76 & 49.77 & 64.29 & 47.92 & 57.14 & 49.49 & 40.48 & 49.08 \\
graph                     & 57.14 & 49.97 & 60.71 & 48.84 & 46.43 & 49.87 & 41.07 & 49.20 \\
bar\_chart                & 53.06 & 49.91 & 69.39 & 46.09 & 51.02 & 49.99 & 40.82 & 49.15 \\
histogram                 & 35.71 & 47.92 & 71.43 & 45.18 & 35.71 & 47.92 & 50.00 & 50.00 \\
venn\_diagram             & 57.14 & 49.49 & 85.71 & 34.99 & 57.14 & 49.49 & 57.14 & 49.49 \\
vector\_plot              & 71.43 & 45.18 & 100.00 & 0.00 & 85.71 & 34.99 & 85.71 & 34.99 \\
other                     & 42.86 & 49.49 & 57.14 & 49.49 & 42.86 & 49.49 & 42.86 & 49.49 \\
line\_chart,bar\_chart    & 28.57 & 45.18 & 71.43 & 45.18 & 14.29 & 34.99 & 28.57 & 45.18 \\
flow\_chart               & 85.71 & 34.99 & 85.71 & 34.99 & 71.43 & 45.18 & 42.86 & 49.49 \\
tree,graph                & 28.57 & 45.18 & 42.86 & 49.49 & 42.86 & 49.49 & 14.29 & 34.99 \\
illustrative\_diagram     & 28.57 & 45.18 & 71.43 & 45.18 & 28.57 & 45.18 & 57.14 & 49.49 \\
line\_chart,scatter\_plot & 71.43 & 45.18 & 71.43 & 45.18 & 42.86 & 49.49 & 42.86 & 49.49 \\
heat\_map                 & 57.14 & 49.49 & 71.43 & 45.18 & 28.57 & 45.18 & 57.14 & 49.49 \\
\bottomrule
\end{tabularx}
\caption{Scores for exact match across models for various chart types. Accuracy and standard deviation (both in \%) are shown.}
\label{tab:ensemble-metrics-1}
\end{table*}

%%% 
%\begin{table*}[ht]
%\centering
%\renewcommand{\arraystretch}{1.0}
%\begin{tabularx}{\textwidth}{p{3.5cm} *{8}{>{\centering\arraybackslash}X}}
%\toprule
%\textbf{QA Pair Type} & \multicolumn{2}{c}{\textbf{Bespoke}} & \multicolumn{2}{c}{\textbf{InternVL3}} & \multicolumn{2}{c}{\textbf{Qwen2.5 VL}} & \multicolumn{2}{c}{\textbf{Phi-4}} \\
%& Acc. (mean) & Std. Dev. & Acc. (mean) & Std. Dev. & Acc. (mean) & Std. Dev. & Acc. (mean) & Std. Dev. \\
%\midrule
%closed-ended infinite answer set visual           & 22.08 & 41.48 & 33.75 & 47.29 & 22.92 & 42.03 & 9.58  & 29.44 \\
%closed-ended infinite answer set non-visual       & 49.58 & 50.00 & 54.58 & 49.79 & 50.42 & 50.00 & 17.92 & 38.35 \\
%closed-ended finite answer set binary visual      & 60.83 & 48.81 & 67.92 & 46.68 & 55.83 & 49.66 & 52.08 & 49.96 \\
%closed-ended finite answer set binary non-visual  & 66.25 & 47.29 & 76.25 & 42.56 & 63.75 & 48.07 & 50.83 & 49.99 \\
%closed-ended finite answer set non-binary visual  & 48.33 & 49.97 & 67.92 & 46.68 & 38.75 & 48.72 & 36.25 & 48.07 \\
%closed-ended finite answer set non-binary non-visual & 40.00 & 48.99 & 57.92 & 49.37 & 30.42 & 46.01 & 26.25 & 44.00 \\
%\bottomrule
%\end{tabularx}
%\caption{Scores for exact match across models for various QA pair types. Accuracy and standard deviation (both in \%) are shown.}
%\label{tab:ensemble-metrics-2}
%\end{table*}
%%%

% \FloatBarrier
\begin{table*}
% [!htb]
\centering
\renewcommand{\arraystretch}{1.0}
% \begin{tabularx}{\textwidth}{p{0.7cm}XXXXXXXXXX}
\begin{tabularx}{\textwidth}{p{0.7cm}p{2.5cm}*{9}{>{\centering\arraybackslash}X}}
\toprule
\textbf{\makebox[1.2em][c]{Rank}} & \textbf{Team} & \textbf{\makebox[2.2em][c]{R1-F1}} & \textbf{R1-P} & \textbf{R1-R} & \textbf{\makebox[2.2em][c]{RL-F1}} & \textbf{RL-P} & \textbf{\makebox[2.2em][c]{RL-R}} & \textbf{\makebox[2.2em][c]{BS-F1}} & \textbf{BS-P} & \textbf{BS-R} \\
\midrule
1 & ExpertNeurons  & 0.805 & 0.809 & 0.811 & 0.804 & 0.808 & 0.810 & 0.985 & 0.985 & 0.985 \\
2 & THAii\_LAB & 0.790 & 0.796 & 0.795 & 0.789 & 0.795 & 0.794 & 0.984 & 0.984 & 0.984 \\
3 & Coling\_UniA & 0.786 & 0.798 & 0.786 & 0.786 & 0.796 & 0.785 & 0.982 & 0.983 & 0.981 \\
4 & florian & 0.763 & 0.766 & 0.770 & 0.762 & 0.765 & 0.769 & 0.983 & 0.983 & 0.984 \\
\underline{5} & \underline{Infyn} & \underline{0.735} & \underline{0.744} & \underline{0.744} & \underline{0.734} & \underline{0.743} & \underline{0.743} & \underline{0.979} & \underline{0.978} & \underline{0.980} \\
6 & Soham Chitnis & 0.706 & 0.719 & 0.705 & 0.705 & 0.719 & 0.704 & 0.980 & 0.982 & 0.979 \\
7 & psr123 & 0.607 & 0.609 & 0.617 & 0.606 & 0.608 & 0.616 & 0.959 & 0.959 & 0.959 \\
\bottomrule
\end{tabularx}
\caption{Top-7 leaderboard rankings}
\label{tab:leaderboard}
\end{table*}
% \FloatBarrier

\end{document}